%% file: main.tex
\documentclass[10pt,twocolumn,letterpaper]{article}

\usepackage[pagenumbers]{cvpr}   

\input{preamble}

\definecolor{cvprblue}{rgb}{0.21,0.49,0.74}
\usepackage[breaklinks,colorlinks,allcolors=cvprblue]{hyperref}

\def\confName{CVPR}
\def\confYear{2027}

\title{Beyond the Camera: A Tri-Modal Driving World Model for Future RGB, Thermal, and LiDAR Forecasting in Adverse Weather}

\author{%
\parbox{\textwidth}{\centering
Faysal Mehrab Chowdhury$^{1}$,\;
Md. Mahfuzur Rahman$^{2}$,\;
Ahmed Rafi Hasan$^{3}$,\;
Kishor Datta Gupta$^{2}$,\;
Mohd Ariful Haque$^{2}$,\;
Savannah Shannon$^{2}$\\[5pt]
\small
$^{1}$Dept. of Computer Science and Engineering, North South University, Dhaka, Bangladesh\\
$^{2}$Dept. of Cyber Physical Systems, Clark Atlanta University, Atlanta, GA, USA\\
$^{3}$Dept. of Computer Science and Engineering, United International University, Dhaka, Bangladesh\\[5pt]
\footnotesize\tt
faysal.chowdhury4@northsouth.edu,\;
mdmahfuzur.rahman@students.cau.edu,\;
ahasan191131@bscse.uiu.ac.bd,\;
kgupta@cau.edu,\;
mohdariful.haque@students.cau.edu,\;
savannah.shannon@students.cau.edu}
}

\begin{document}
\maketitle
\input{sec/0_abstract}
\input{sec/1_intro}
\input{sec/2_related_work}
\input{sec/3_methodology}
\input{sec/4_experimentation}
\input{sec/5_results_analysis}
\input{sec/6_conclusion}
\input{sec/acknowledgement}
{
    \small
    \bibliographystyle{ieeenat_fullname}
    \bibliography{main}
}

\clearpage
\input{sec/X_suppl}

\end{document}

%% file: preamble.tex
\usepackage{float}



%% file: sec/0_abstract.tex
\begin{abstract}
Driving world models forecast what a vehicle's sensors will record next. Most forecast camera video and some add LiDAR, but thermal cameras, which still see in darkness and some adverse weather, are rarely a forecast target, and to our knowledge no world model forecasts all three sensors.
We study joint forecasting of RGB, thermal, and LiDAR on real adverse-weather sequences of the DSERT-RoLL dataset.
We post-train a pretrained video world model for each sensor and represent LiDAR as a depth video aligned with the RGB camera, with densification and encoding chosen by what the model's video tokenizer can reconstruct.
A tri-sensor model sees all three sensors as one video.
We score LiDAR in meters against measured returns and compare against persistence baselines passed through the same tokenizer as the model. With this matching, an apparent loss of the thermal model to persistence in fog disappears.
On all 58 validation sequences in six weather conditions, post-trained RGB and thermal forecasts beat compression-matched persistence.
Joint training lowers the LiDAR error at 2\,s by 2.1\,m at a small cost in camera fidelity, and only the tri-sensor model's LiDAR forecasts beat persistence in mean error.
Blanking one sensor degrades the forecasts of the others more in snow than in fog.
We release the post-trained tri-sensor model, TriCosmos-2.5, at \url{https://huggingface.co/SORA-Lab/TriCosmos-2.5-predict}.
\end{abstract}

%% file: sec/1_intro.tex
\section{Introduction}
\label{sec:intro}

\begin{figure*}[t]
  \centering
  \includegraphics[width=\textwidth]{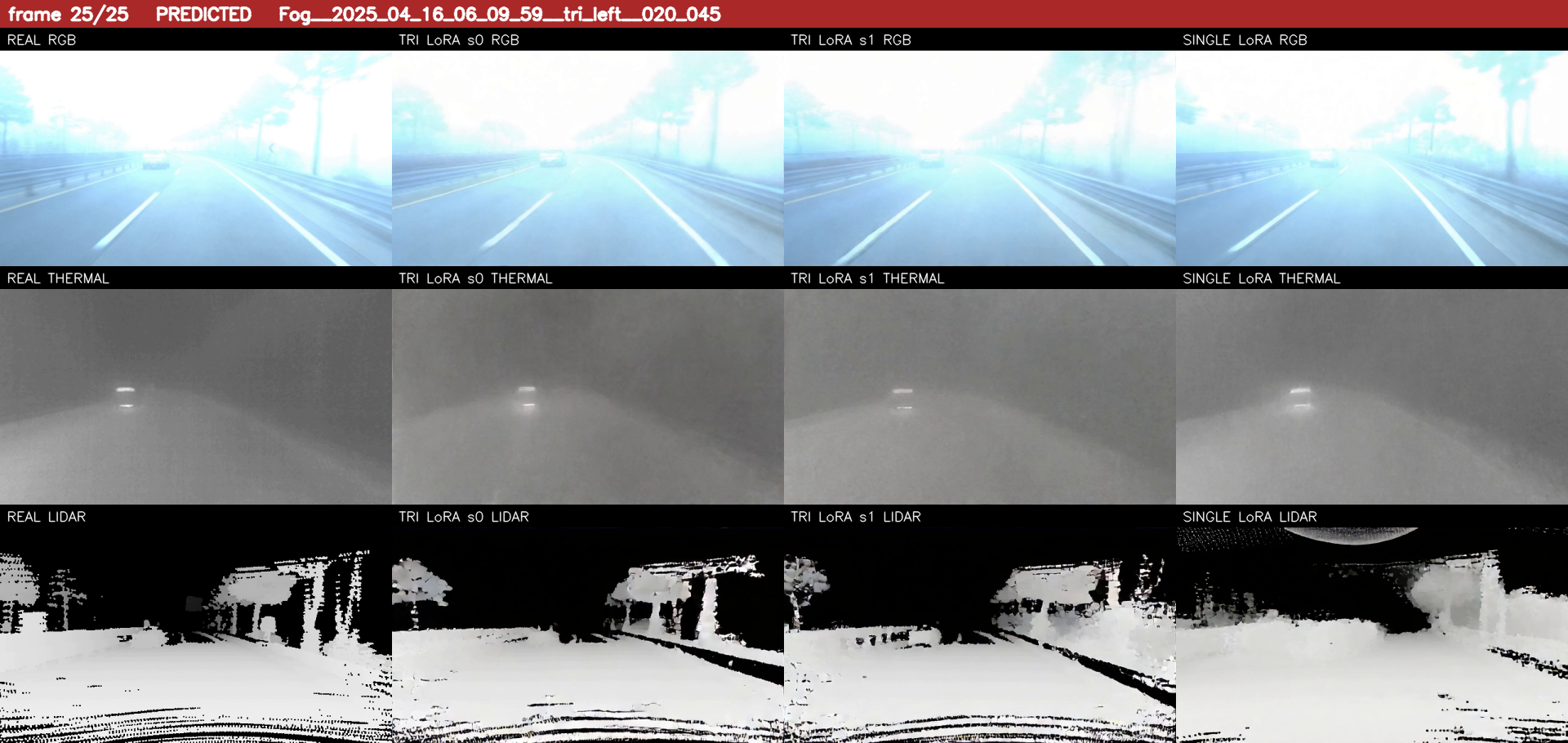}
  \caption{\textbf{Forecasting beyond the camera.} From five observed frames, our models forecast the next 2\,s of RGB, thermal, and LiDAR depth in real adverse weather. Shown: the last forecast frame (2\,s) of one fog window. Columns: real future, two samples of the tri-sensor model, and the single-sensor model. The tri-sensor LiDAR forecasts stay close to the real depth, while the single-sensor LiDAR forecast is distorted. The camera forecasts of the two models look similar; in all of them, the vehicle on the left differs from the real future and between samples.}
  \label{fig:teaser}
\end{figure*}

A driving world model forecasts the frames that a vehicle's sensors will record next, given the frames recorded so far. Such forecasts serve to generate training data, to test planners, and to anticipate other road users~\cite{hu2023gaia1,wang2024drivedreamer,nvidia2025cosmos}. Most driving world models forecast camera video, and recent ones also generate LiDAR~\cite{guo2025genesis,xie2025xdrive,nvidia2026cosmos3}. Camera images, however, lose much of their contrast in fog, snow, and darkness~\cite{bijelic2020seeing}. A thermal camera measures emitted heat instead of reflected light, and a LiDAR measures distance directly, so both can retain information that the camera loses. To our knowledge, no world model forecasts RGB, thermal, and LiDAR observations together, and no study compares how well these sensors can be forecast in real adverse weather.

World models range from general to driving-specific. Dreamer~\cite{hafner2020dreamer} learns a latent world model for control from images, and Atlas~\cite{worldlabs2026atlas} generates camera-controlled video and 3D scenes from images. Among driving world models, DrivingWorld~\cite{hu2024drivingworld} forecasts front-camera video from about 3{,}456 hours of driving, DriveDreamer~\cite{wang2024drivedreamer} is trained on nuScenes~\cite{caesar2020nuscenes}, and Vista~\cite{gao2024vista} on public driving videos and nuScenes. Copilot4D~\cite{zhang2024copilot4d} forecasts LiDAR on nuScenes, KITTI, and Argoverse~2, and Genesis~\cite{guo2025genesis} generates camera video and LiDAR on nuScenes. To our knowledge, none of these models forecasts thermal observations, and none is evaluated separately on real fog and snow.

Adding these sensors to a video world model raises two problems. The first is representation. Video world models are pretrained on RGB video, and a LiDAR sweep projected into a camera frame covers about 3\% of its pixels. We find that the model's video tokenizer cannot reconstruct such sparse input (\cref{sec:lidar}). The second is evaluation. A persistence baseline that repeats the last observed frame is hard to beat when much of the scene is static, and every forecast passes through a lossy tokenizer that the baseline never sees. For the thermal camera in fog, this compression gap alone decides whether the model looks worse than persistence or no different from it (\cref{sec:fair}).

We study future forecasting of RGB, thermal, and LiDAR on the real adverse-weather sequences of DSERT-RoLL~\cite{dsertroll2026}. DSERT-RoLL was released with 2D and 3D object detection benchmarks~\cite{dsertroll2026}; to our knowledge, it has not been used to forecast future observations of these three sensors. We post-train the video foundation model Cosmos-Predict2.5~\cite{nvidia2025predict25} for each sensor with low-rank adaptation~\cite{hu2022lora}; without post-training, it forecasts all three sensors worse (\cref{sec:zeroshot}). LiDAR enters the model as a depth video aligned with the RGB camera, built from ego-motion-compensated sweeps, with densification and encoding chosen by what the tokenizer can reconstruct. A tri-sensor model then sees all three sensors as one video, and we measure how much each forecast relies on the other sensors by blanking one sensor at a time.

Our contributions are:
\begin{itemize}
  \item We study the joint forecasting of RGB, thermal, and LiDAR observations in six real weather conditions, including fog, rain, and snow, with one post-trained model per sensor and one for all three sensors (\cref{sec:method}).
  \item We design a camera-aligned LiDAR depth video whose densification and encoding are tested against the world model's tokenizer. On one Clear sequence, adding morphological closing to dilation lowers the median near-range error after the tokenizer from \(1.57\)\,m to \(0.11\)\,m (\cref{sec:lidar}).
  \item We propose an evaluation protocol shared by all three sensors, with sensor-specific metrics: metric depth for LiDAR, object-level checks from the dataset's 2D and 3D boxes, persistence baselines matched to the model's compression, and confidence intervals over sequences. Without compression matching, the thermal model in fog looks worse than persistence; with it, there is no clear difference (\cref{sec:protocol,sec:fair}).
  \item On all 58 validation sequences, we show that the tri-sensor model lowers the LiDAR error by 2.1\,m compared with the single-sensor model (4.68\,m vs.\ 6.81\,m) and makes LiDAR forecasts better than compression-matched persistence in every weather condition with at least three sequences, except at near range. Its camera forecasts stay close to those of the single-sensor models, with a small loss in LPIPS. The model relies more on the other sensors in snow than in fog, but within one weather condition its reliance does not follow measured sensor quality (\cref{sec:joint,sec:reliance}).
\end{itemize}

%% file: sec/2_related_work.tex
\section{Related work}
\label{sec:related}

\paragraph{Driving world models.}
Video world models for driving forecast future camera frames from past frames, text, or actions. GAIA-1~\cite{hu2023gaia1} and DriveDreamer~\cite{wang2024drivedreamer} condition video generation on driving context, and video foundation models such as Cosmos~\cite{nvidia2025cosmos,nvidia2025predict25} are post-trained for driving. These models forecast RGB video. We build on the same type of video foundation model and add thermal video and LiDAR depth as forecast targets.

\paragraph{Multi-sensor and LiDAR generation.}
Genesis~\cite{guo2025genesis} couples a video branch with a LiDAR branch, X-Drive~\cite{xie2025xdrive} synthesizes camera images and LiDAR from shared layouts, and Cosmos~3~\cite{nvidia2026cosmos3} and M\(^4\)World~\cite{m4world2026} generate synchronized multi-view video and LiDAR. LiDAR world models forecast future point clouds~\cite{zhang2024copilot4d,lidarverse2026}, and Cosmos-Drive-Dreams~\cite{nvidia2025drivedreams} uses LiDAR depth video as a control signal. These methods pair cameras with LiDAR; none of them forecasts thermal observations. We forecast all three sensors and evaluate in real fog, rain, and snow.

\paragraph{Thermal image generation.}
Most thermal synthesis methods translate single RGB images into thermal images~\cite{thermalgen2025,thera2026,diffv2ir2025}. Thermal video also serves as input for forecasting agent trajectories at night~\cite{thermalforecast2025}, where the output is a set of trajectories. Li \etal~\cite{li2020nightscene} predict future infrared frames of night driving scenes with a recurrent network. We forecast thermal frames with a video world model, with and without the other two sensors.

\paragraph{Adverse weather.}
Bijelic \etal~\cite{bijelic2020seeing} fuse several sensors, including camera, LiDAR, and radar, for detection in fog and snow, and recent generative methods change the weather of camera videos~\cite{realweather2025,meteoverse2026}. DSERT-RoLL~\cite{dsertroll2026} provides synchronized RGB, thermal, event, radar, and LiDAR data in six weather conditions, with 2D and 3D annotations and perception benchmarks. We use it for forecasting.

\paragraph{Evaluating forecasts.}
Video generation is commonly scored with image metrics such as PSNR, SSIM~\cite{wang2004ssim}, and LPIPS~\cite{zhang2018lpips}, and with distribution metrics such as FVD~\cite{unterthiner2018fvd} and FID~\cite{heusel2017fid}. WorldLens~\cite{worldlens2026} evaluates driving world models beyond image metrics. Repeating the last observed frame is a strong baseline when much of the scene is static. We match this baseline to the model's tokenizer, add metric-depth and object-level scores, and report confidence intervals over sequences.

%% file: sec/3_methodology.tex
\section{Method}
\label{sec:method}

\begin{figure*}[t]
  \centering
  \includegraphics[width=\linewidth]{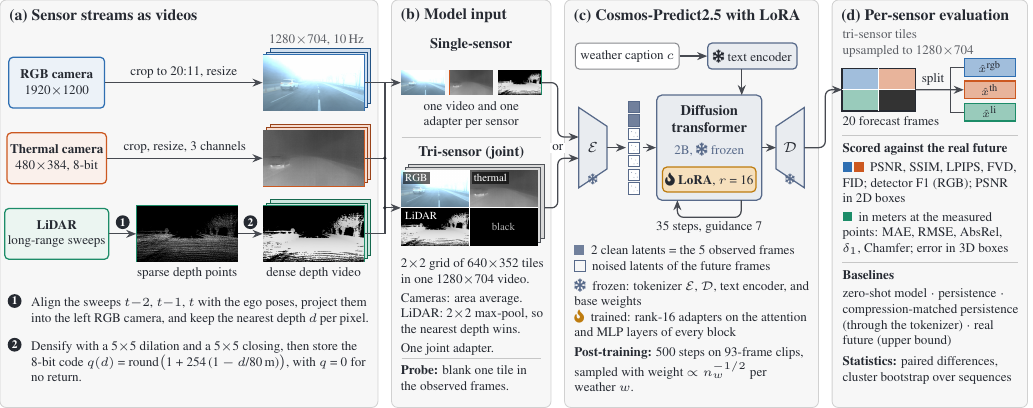}
  \caption{\textbf{Overview.} (a) RGB and thermal frames are cropped and resized to \(1280\times704\) videos; LiDAR sweeps become a depth video aligned with the RGB camera. (b) The single-sensor models forecast each video on its own; the tri-sensor model forecasts a \(2\times2\) grid of all three sensors, with the fourth tile left black. (c) Cosmos-Predict2.5 with low-rank adapters forecasts 20 frames from 5 observed frames. (d) Tri-sensor forecasts are split back into their sensors, and every forecast is scored against the real future.}
  \label{fig:pipeline}
\end{figure*}

\paragraph{Problem setup.}
For each sensor \(s \in \{\mathrm{rgb}, \mathrm{th}, \mathrm{li}\}\) a window is a sequence of frames \(x^s_{1:C+H}\) at 10\,Hz. Given the first \(C=5\) observed frames, a world model \(f\) produces a forecast \(\hat{x}^s_{C+1:C+H} = f(x^s_{1:C}, c)\) of the next \(H=20\) frames (2\,s), where \(c\) is a text caption. The single-sensor setting uses one model per sensor; the joint setting conditions every sensor's forecast on all three, \(\hat{x}^{\mathrm{rgb,th,li}}_{C+1:C+H} = f(x^{\mathrm{rgb,th,li}}_{1:C}, c)\). \Cref{fig:pipeline} gives an overview.

\subsection{Sensor-specific post-training}
\label{sec:posttrain}
We start from the 2B-parameter video-to-world model of Cosmos-Predict2.5~\cite{nvidia2025predict25}, a latent diffusion transformer that operates on the latents of a video tokenizer at \(1280\times704\) resolution. Its text encoder and tokenizer stay frozen. For each sensor we train low-rank adapters~\cite{hu2022lora} (rank 16, \(\alpha=32\)) on the attention projections (query, key, value, output) and both MLP layers of every transformer block, with the base weights frozen. All runs share one recipe: learning rate \(10^{-4}\), weight decay \(10^{-3}\), 50 linear warm-up steps followed by linear decay to 10\% over 500 optimizer steps, an effective batch of 4 clips of 93 frames, and bfloat16 precision. Clips are sampled with weights proportional to \(n_w^{-1/2}\), where \(n_w\) is the number of training clips of their weather condition, so rare conditions are seen more often without dominating. Training conditions on the first two latent frames, which correspond to the five observed frames. Captions are generated from the sequence labels: for RGB and thermal they name the weather and lighting, for LiDAR they read ``front lidar depth video, \textit{weather}''. Thermal frames are stored as 8-bit gray images replicated to three channels.

\subsection{LiDAR as a camera-aligned depth video}
\label{sec:lidar}
A video model expects dense images, while a LiDAR sweep is a sparse set of 3D points. We render LiDAR into the image plane of the left RGB camera so that the model sees a depth video with the same geometry as the RGB stream.

\paragraph{Accumulation and projection.}
At frame \(t\) we accumulate the sweeps \(t-2\), \(t-1\), and \(t\) of the long-range LiDAR, transform each into the sensor frame of \(t\) with the ego poses provided by the dataset, and project the points into the rectified left RGB camera with the dataset calibration. Each pixel keeps the nearest depth \(d\). Ego-motion compensation uses the dataset poses in the LiDAR frame; a depth-agreement test against the camera image selects this convention as best or equivalent on 42 of 46 moving sequences, and IMU-derived alternatives are worse.

\paragraph{Densification.}
Projected points cover about 3\% of the frame, and the tokenizer cannot reconstruct such sparse input (\cref{tab:lidar_design}). We fill gaps with a \(5\times5\) elliptical dilation followed by a \(5\times5\) morphological closing. Dilation alone leaves holes that the tokenizer blurs into wrong depths; on one Clear sequence, the closing step removes them and cuts the median near-range error after the tokenizer from 1.57\,m to 0.11\,m.

\paragraph{Encoding.}
Depth up to \(d_{\max}=80\)\,m is stored as an 8-bit code
\begin{equation}
  q(d) = \operatorname{round}\!\left(1 + 254\left(1 - \frac{d}{d_{\max}}\right)\right),
  \label{eq:depthcode}
\end{equation}
with \(q=0\) for pixels without a return, replicated to three channels, so near objects are bright and empty pixels black. Rounding instead of truncation lowers the quantization error floor from about 0.15\,m to 0.08\,m. We compared this linear code with logarithmic and inverse-depth codes through the tokenizer (\cref{tab:lidar_design}b). The alternatives reduce near-range error but increase overall and far-range error, and the tokenizer loses about 0.5\,m at near range while forecasting errors are an order of magnitude larger (\cref{sec:experiments}), so we keep the linear code. We also tested a filter that removes snow returns and rejected it, since it increased the error in every weather condition. The field of view of the long-range LiDAR is wider than the camera's, so the depth video covers only the camera-visible part of each sweep.

\begin{table}[t]
  \centering
  \small
  \setlength{\tabcolsep}{3pt}
  \caption{\textbf{The LiDAR depth video is designed against the tokenizer.} PSNR in dB. (a) Without closing, holes survive densification and the tokenizer turns them into wrong depths. Measured on frames 5--29 of one Clear sequence. (b) Logarithmic and inverse-depth codes help near range but hurt overall and far range; near \(<\!10\)\,m, far \(\geq\!30\)\,m. Measured on 4 windows per evaluated weather of the initial split. -- = not measured.}
  \label{tab:lidar_design}
  \begin{tabular}{@{}lcccc@{}}
    \toprule
    \multicolumn{5}{@{}l}{\emph{(a) Densification, one sequence, after the tokenizer}} \\
    \midrule
    Input & PSNR\,\(\uparrow\) & SSIM\,\(\uparrow\) & \multicolumn{2}{c}{Near med. (m)\,\(\downarrow\)} \\
    \midrule
    Raw points & 15.6 & 0.498 & \multicolumn{2}{c}{--} \\
    Dilation & 13.44 & -- & \multicolumn{2}{c}{1.57} \\
    Dilation\,+\,closing (ours) & \textbf{26.32} & -- & \multicolumn{2}{c}{\textbf{0.11}} \\
    \midrule
    \multicolumn{5}{@{}l}{\emph{(b) Depth code, 12 windows, error (m) after the tokenizer}} \\
    \midrule
    Code & Mean\,\(\downarrow\) & Median\,\(\downarrow\) & Near\,\(\downarrow\) & Far\,\(\downarrow\) \\
    \midrule
    Linear (ours) & \textbf{1.61} & \textbf{0.62} & 0.54 & \textbf{4.10} \\
    Logarithmic & 1.71 & 0.67 & 0.32 & 4.67 \\
    Inverse depth & 2.71 & 1.33 & \textbf{0.28} & 7.35 \\
    \bottomrule
  \end{tabular}
\end{table}

\subsection{Joint tri-sensor model}
\label{sec:tri}
The tri-sensor model sees all three sensors as one video. Each \(1280\times704\) sensor frame is reduced to a \(640\times352\) tile and the tiles form a \(2\times2\) grid of the native \(1280\times704\) size: RGB top left, thermal top right, LiDAR bottom left, and the fourth tile black. RGB and thermal tiles are area-averaged. The LiDAR tile is max-pooled over \(2\times2\) blocks, because averaging depth codes with empty pixels would create depths that no point measured; the nearest depth wins, as in the densification. Windows are matched across the three sensors by sequence and frame range, so every grid shows the same moment. The thermal tile shows the thermal camera's own view; it is not warped to the RGB image. The tri-sensor model is trained with the same recipe as the single-sensor models (\cref{sec:posttrain}), with a caption that names the layout and the weather and lighting. Every forecast grid is split back into its tiles, which are upsampled to \(1280\times704\) (bilinear for cameras, nearest for LiDAR) and scored by the unchanged single-sensor evaluation.

%% file: sec/4_experimentation.tex
\section{Experiments}
\label{sec:experiments}

\subsection{Data and protocol}
\label{sec:protocol}

\paragraph{Data.}
DSERT-RoLL~\cite{dsertroll2026} contains 190 sequences in six weather conditions (Clear, Fog, Light and Heavy Rain, Light and Heavy Snow), with an official split of 132 training and 58 validation sequences. We use the left RGB camera, the left thermal camera, and the long-range LiDAR. Our models are trained on 35 of the official training sequences (Clear 5, Fog 23, Heavy Rain 1, Heavy Snow 5, Light Rain 1); Light Snow does not occur in training. Training clips are 93-frame windows, the clip length of the pretrained model, with a stride of 8 frames. Evaluation uses all 58 official validation sequences. An evaluation window has 25 frames (5 observed, 20 forecast) with a stride of 10 frames, and the same windows are used for every sensor and model. The evaluation covers 300 windows: Clear 174 (30 sequences), Fog 31 (9), Light Rain 31 (5), Heavy Rain 26 (6), Light Snow 25 (6), and Heavy Snow 13 (2). We call all conditions except Clear \emph{adverse}.

\paragraph{Forecasts.}
Every model generates its forecast with 35 sampling steps and classifier-free guidance 7, conditioned on the five observed frames and the caption, with seeds shared across models so that comparisons are paired. The full-validation protocol uses one sample per window. To bound compute on the 45 validation sequences outside the initial split, it scores 3 of every 5 windows of each sequence, chosen by position before any result was seen and applied identically to all 58 sequences. The 13 sequences of the initial split (\cref{sec:analyses}) were generated with two samples per window; the full-validation results use the first. On the initial split, where the full protocol is available, this choice reverses no verdict (\cref{sec:robust}).

\paragraph{Metrics.}
RGB and thermal frames are 8-bit images (the dataset provides thermal as 8-bit images, not raw radiometric values), and PSNR uses a peak value of 255. For RGB and thermal we report PSNR, SSIM~\cite{wang2004ssim}, and LPIPS~\cite{zhang2018lpips} over the first 10 (1\,s) and 20 (2\,s) forecast frames, and FVD~\cite{unterthiner2018fvd} and FID~\cite{heusel2017fid} as indicative distribution metrics. Object-level agreement for RGB is the F1 score between YOLO11-L detections on the forecast and the dataset's 2D boxes (confidence 0.25, IoU 0.5, boxes of at least 16 pixels). The detector does not transfer to thermal images, so for both cameras we also report PSNR inside the dataset's 2D boxes. LiDAR forecasts are decoded to meters with \cref{eq:depthcode} and compared with the measured points of each future frame up to 75\,m: mean and median absolute error, root mean squared error (RMSE), absolute relative error (AbsRel), the share of points with \(\max(d/\hat d,\, \hat d/d) < 1.25\) (\(\delta_1\)), mean absolute error at near (\(<\!10\)\,m), middle (10--30\,m), and far range (30--75\,m), and the symmetric Chamfer distance between the forecast as 3D points and the measured points. A measured point whose forecast pixel is empty counts as an error at \(d_{\max}\); we report the share of such points separately. Object-level LiDAR error is measured at the points inside the dataset's 3D boxes; the box convention was verified on the point cloud (93\% of each box's points captured).

\paragraph{Baselines.}
\emph{Zero-shot} is the pretrained model with the same captions. \emph{Persistence} repeats the last observed frame. Every forecast passes through the model's tokenizer, while plain persistence does not, so we add \emph{compression-matched persistence}: the last observed frame of the tokenizer's reconstruction of the window, repeated. The real future frames, passed through the same encoding as the forecasts, give the upper bound.

\paragraph{Statistics.}
Windows from one sequence are correlated. We therefore compare models with paired per-window differences and 95\% confidence intervals from a cluster bootstrap over sequences (2000 resamples). A difference is called better or worse only if its interval excludes zero and at least three sequences contribute. Camera results are reported for all windows and for windows where the camera sees a scene; windows whose median frame contrast is below a fixed threshold (RGB 20, thermal 35, on a 0--255 scale) are white-out or black and are marked as not visible.

\subsection{Main results}
\label{sec:main}

\begin{table*}[t]
  \centering
  \small
  \setlength{\tabcolsep}{4.3pt}
  \caption{\textbf{Forecasting on the full DSERT-RoLL validation set} (58 sequences, six weather conditions, 300 windows), 2\,s horizon. All rows are scored at the tile resolution of the tri-sensor model (\cref{sec:joint}). The post-trained single-sensor models beat both persistence baselines on every camera metric. The tri-sensor camera forecasts are close to the single-sensor ones in PSNR and SSIM, but their thermal LPIPS is not better than persistence. For LiDAR, only the tri-sensor model beats persistence in mean error, while persistence stays best at near range. Camera metrics on scene-visible windows (RGB 285, thermal 280). Bold = best, underline = second best, excluding the real future. -- = not applicable or not measured.}
  \label{tab:main}
  \begin{tabular}{@{}l ccc ccc cccc@{}}
    \toprule
    & \multicolumn{3}{c}{RGB (scene visible)} & \multicolumn{3}{c}{Thermal (scene visible)} & \multicolumn{4}{c}{LiDAR} \\
    \cmidrule(lr){2-4}\cmidrule(lr){5-7}\cmidrule(l){8-11}
    Method & PSNR\,\(\uparrow\) & LPIPS\,\(\downarrow\) & F1\,\(\uparrow\) & PSNR\,\(\uparrow\) & SSIM\,\(\uparrow\) & LPIPS\,\(\downarrow\) & MAE (m)\,\(\downarrow\) & \(\delta_1\)\,\(\uparrow\) & Near (m)\,\(\downarrow\) & Obj. (m)\,\(\downarrow\) \\
    \midrule
    Persistence & 19.26 & 0.351 & 0.293 & 22.90 & 0.784 & \underline{0.259} & 5.71 & \underline{0.777} & \textbf{2.99} & 10.32 \\
    Persistence, compr.-matched & 19.27 & 0.362 & 0.301 & 22.89 & 0.787 & 0.272 & \underline{5.71} & 0.755 & \underline{3.03} & 10.25 \\
    Single-sensor LoRA & \textbf{21.79} & \textbf{0.284} & \textbf{0.447} & \textbf{24.86} & \textbf{0.811} & \textbf{0.243} & 6.81 & 0.752 & 7.12 & \underline{8.22} \\
    Tri-sensor LoRA & \underline{21.71} & \underline{0.304} & \underline{0.410} & \underline{24.61} & \underline{0.810} & 0.270 & \textbf{4.68} & \textbf{0.794} & 3.70 & \textbf{7.21} \\
    \midrule
    Real future (upper bound) & 37.56 & 0.102 & 0.607 & 44.25 & 0.980 & 0.055 & 1.25 & 0.944 & 0.31 & -- \\
    \bottomrule
  \end{tabular}
\end{table*}

\begin{table*}[t]
  \centering
  \small
  \setlength{\tabcolsep}{3.4pt}
  \caption{\textbf{Gains over compression-matched persistence by weather condition.} Paired difference (forecast minus compression-matched persistence), 2\,s horizon, 95\% cluster-bootstrap intervals over sequences. Single-sensor models at full resolution; the tri-sensor row is compared with persistence at tile resolution. Number of sequences in parentheses (L.\ = Light, H.\ = Heavy). Camera metrics on scene-visible windows; Heavy Snow has no scene-visible camera window. \textbf{Bold} = better, \(^\ast\) = worse, plain = no clear difference. Heavy Snow has two sequences, so no verdict is given.}
  \label{tab:weather}
  \begin{tabular}{@{}ll cccccc cc@{}}
    \toprule
    Model & Metric & Clear (30) & Fog (9) & L.\ Rain (5) & H.\ Rain (6) & L.\ Snow (6) & H.\ Snow (2) & Adverse (28) & All (58) \\
    \midrule
    RGB & PSNR (dB)\,\(\uparrow\) & \textbf{+2.94} & \textbf{+2.99} & \textbf{+2.27} & \(-\)0.17 & +0.85 & -- & \textbf{+1.60} & \textbf{+2.42} \\
        & LPIPS\,\(\downarrow\) & \textbf{\(-\)0.097} & \textbf{\(-\)0.052} & \textbf{\(-\)0.085} & \textbf{\(-\)0.046} & \textbf{\(-\)0.034} & -- & \textbf{\(-\)0.056} & \textbf{\(-\)0.081} \\
    Thermal & PSNR (dB)\,\(\uparrow\) & \textbf{+2.74} & \(-\)0.97 & +1.26 & +0.04 & \textbf{+2.01} & -- & +0.57 & \textbf{+1.92} \\
        & LPIPS\,\(\downarrow\) & \textbf{\(-\)0.048} & +0.001 & \(-\)0.002 & +0.008 & \textbf{\(-\)0.012} & -- & \(-\)0.002 & \textbf{\(-\)0.031} \\
    LiDAR & MAE (m)\,\(\downarrow\) & +0.90 & +4.85\(^\ast\) & \(-\)0.61 & \(-\)0.35 & \(-\)0.69 & +5.65 & +1.42 & +1.11\(^\ast\) \\
    Tri-sensor LiDAR & MAE (m)\,\(\downarrow\) & \textbf{\(-\)0.68} & \textbf{\(-\)1.46} & \textbf{\(-\)1.46} & \textbf{\(-\)1.29} & \textbf{\(-\)1.34} & \(-\)2.43 & \textbf{\(-\)1.50} & \textbf{\(-\)1.03} \\
    \bottomrule
  \end{tabular}
\end{table*}

\Cref{tab:main} compares all models on the full validation set, and \cref{tab:weather} breaks the main comparison down by weather condition.

\paragraph{Cameras.}
Post-training makes both camera forecasts better than compression-matched persistence. For RGB on scene-visible windows, PSNR at 2\,s rises by 2.42\,dB (\([+2.01, +2.82]\)), LPIPS falls by 0.081 (\([-0.093, -0.070]\)), and the F1 score of detected objects rises by 0.157 (\([+0.113, +0.199]\)). LPIPS improves in every weather condition with scene-visible windows; PSNR improves in Clear, Fog, and Light Rain and shows no clear difference in Heavy Rain and Light Snow. For thermal, PSNR rises by 1.92\,dB (\([+1.45, +2.33]\)) and LPIPS falls by 0.031 (\([-0.039, -0.022]\)), but the gain comes from Clear and Light Snow. In Fog and rain, the thermal model is not clearly different from compression-matched persistence.

\paragraph{LiDAR.}
The single-sensor LiDAR model does not beat persistence on the full validation set. Its mean error at 2\,s is 1.11\,m higher than that of compression-matched persistence (\([+0.04, +2.35]\)), and the gap comes from near range (\(+4.16\)\,m, \([+2.58, +6.20]\)) and from Fog (\(+4.85\)\,m, \([+1.79, +7.63]\)). At far range it is better (\(-1.38\)\,m, \([-2.15, -0.47]\)). The tri-sensor model is better than compression-matched persistence in mean error (\(-1.03\)\,m, \([-1.36, -0.72]\)) and in every weather condition with at least three sequences (\cref{tab:weather}). It is also better in RMSE (\(-1.86\)\,m), AbsRel (\(-0.024\)), \(\delta_1\) (\(+0.039\)), middle range (\(-1.26\)\,m), far range (\(-2.74\)\,m), and inside 3D boxes (\(-3.03\)\,m, \([-3.87, -2.27]\)), but worse at near range (\(+0.67\)\,m, \([+0.36, +0.95]\)). The supplement lists AbsRel, RMSE, and the middle- and far-range errors of every model.

\subsection{Joint versus separate forecasting}
\label{sec:joint}
The tri-sensor model sees each sensor at half resolution. For a fair comparison, the single-sensor forecasts go through the same reduction and upsampling as a tri-sensor tile, so the two differ only in whether the sensors were forecast jointly. The single-sensor models were trained at full resolution, so this comparison favors them. On the full validation set, the tri-sensor model lowers the LiDAR error at 2\,s by 2.13\,m compared with the single-sensor model (4.68\,m vs.\ 6.81\,m) and improves every LiDAR metric we measure (\cref{tab:joint}). The gain is largest in Fog (\(-6.32\)\,m, \([-8.57, -3.65]\)) and is also significant in Clear (\(-1.56\)\,m); for Light Rain, Heavy Rain, and Light Snow taken separately the intervals include zero. Both cameras lose a little fidelity: LPIPS rises by 0.020 for RGB and 0.027 for thermal, fewer objects are detected in the RGB forecasts (F1 \(-0.041\)), and thermal loses 0.25\,dB PSNR. On the initial split, FID rises from 53.9 to 123.8 for thermal; for RGB, FVD rises from 324 to 428 while FID falls from 70.5 to 57.4. These distribution metrics rest on 90 windows and are indicative only.

\begin{table}[t]
  \centering
  \small
  \setlength{\tabcolsep}{3.5pt}
  \caption{\textbf{Joint training helps LiDAR; the camera forecasts change little.} Paired differences (tri-sensor minus single-sensor), both at tile resolution, 2\,s horizon, full validation set, 95\% cluster-bootstrap intervals over sequences; camera metrics on scene-visible windows. All intervals exclude zero except for RGB PSNR.}
  \label{tab:joint}
  \begin{tabular}{@{}llcc@{}}
    \toprule
    Sensor & Metric & Tri \(-\) single & 95\% CI \\
    \midrule
    LiDAR & MAE (m)\,\(\downarrow\) & \(-2.13\) & \([-3.21, -1.25]\) \\
          & RMSE (m)\,\(\downarrow\) & \(-3.15\) & \([-4.47, -2.02]\) \\
          & AbsRel\,\(\downarrow\) & \(-0.223\) & \([-0.341, -0.128]\) \\
          & Near MAE (m)\,\(\downarrow\) & \(-3.42\) & \([-5.35, -1.90]\) \\
          & \(\delta_1\)\,\(\uparrow\) & \(+0.043\) & \([+0.017, +0.071]\) \\
          & Chamfer (m)\,\(\downarrow\) & \(-0.13\) & \([-0.30, -0.002]\) \\
          & Obj. MAE (m)\,\(\downarrow\) & \(-1.01\) & \([-1.83, -0.26]\) \\
    \midrule
    RGB & PSNR (dB)\,\(\uparrow\) & \(-0.09\) & \([-0.40, +0.25]\) \\
        & LPIPS\,\(\downarrow\) & \(+0.020\) & \([+0.014, +0.026]\) \\
        & F1\,\(\uparrow\) & \(-0.041\) & \([-0.066, -0.017]\) \\
    \midrule
    Thermal & PSNR (dB)\,\(\uparrow\) & \(-0.25\) & \([-0.49, -0.02]\) \\
            & LPIPS\,\(\downarrow\) & \(+0.027\) & \([+0.022, +0.032]\) \\
    \bottomrule
  \end{tabular}
\end{table}

%% file: sec/5_results_analysis.tex
\section{Analysis}
\label{sec:analysis}

\subsection{Where the forecasts gain and where they do not}
\label{sec:where}
\Cref{tab:weather} shows that the gains over compression-matched persistence depend on both the sensor and the weather condition.

\paragraph{Camera gains depend on the weather.}
The RGB model is better in PSNR by 2.94\,dB in Clear and 2.99\,dB in Fog, and its LPIPS is lower in every condition with scene-visible windows. In Heavy Rain and Light Snow its PSNR is not clearly different from persistence. The thermal model gains 2.74\,dB in Clear and 2.01\,dB in Light Snow, but shows no clear difference in Fog, Light Rain, and Heavy Rain. Fog makes up 23 of the 35 training sequences and Clear only 5, so the size of the gain does not follow the amount of training data for a condition.

\begin{figure}[t]
  \centering
  \setlength{\tabcolsep}{0.5pt}%
  \newlength{\lsw}\setlength{\lsw}{\dimexpr(\columnwidth-14pt)/3\relax}%
  \newlength{\lsh}\setlength{\lsh}{0.5523\lsw}
  \def\lsimg#1#2#3#4#5{\includegraphics[width=\lsw,trim=#4 #2 #5 #3,clip]{fig/Frame_Comparison_fog/Light_snow/light_snow_#1.png}}%
  \def\lslab#1{\rotatebox{90}{\makebox[\lsh][c]{\scriptsize #1}}}%
  \def\lsrow#1#2#3#4{\lslab{#4} & \lsimg{#1}{#2}{#3}{1}{1441} & \lsimg{#1}{#2}{#3}{481}{961} & \lsimg{#1}{#2}{#3}{1441}{1}}%
  \def\lspanel#1#2{& \multicolumn{3}{c}{\footnotesize #2} \\
    \lsrow{#1}{610}{67}{RGB} \\[-2pt]
    \lsrow{#1}{318}{359}{Thermal} \\[-2pt]
    \lsrow{#1}{26}{651}{LiDAR} \\}%
  \begin{tabular}{@{}c@{\hspace{3pt}}ccc@{}}
    & \scriptsize Real & \scriptsize Tri-sensor & \scriptsize Single-sensor \\
    \lspanel{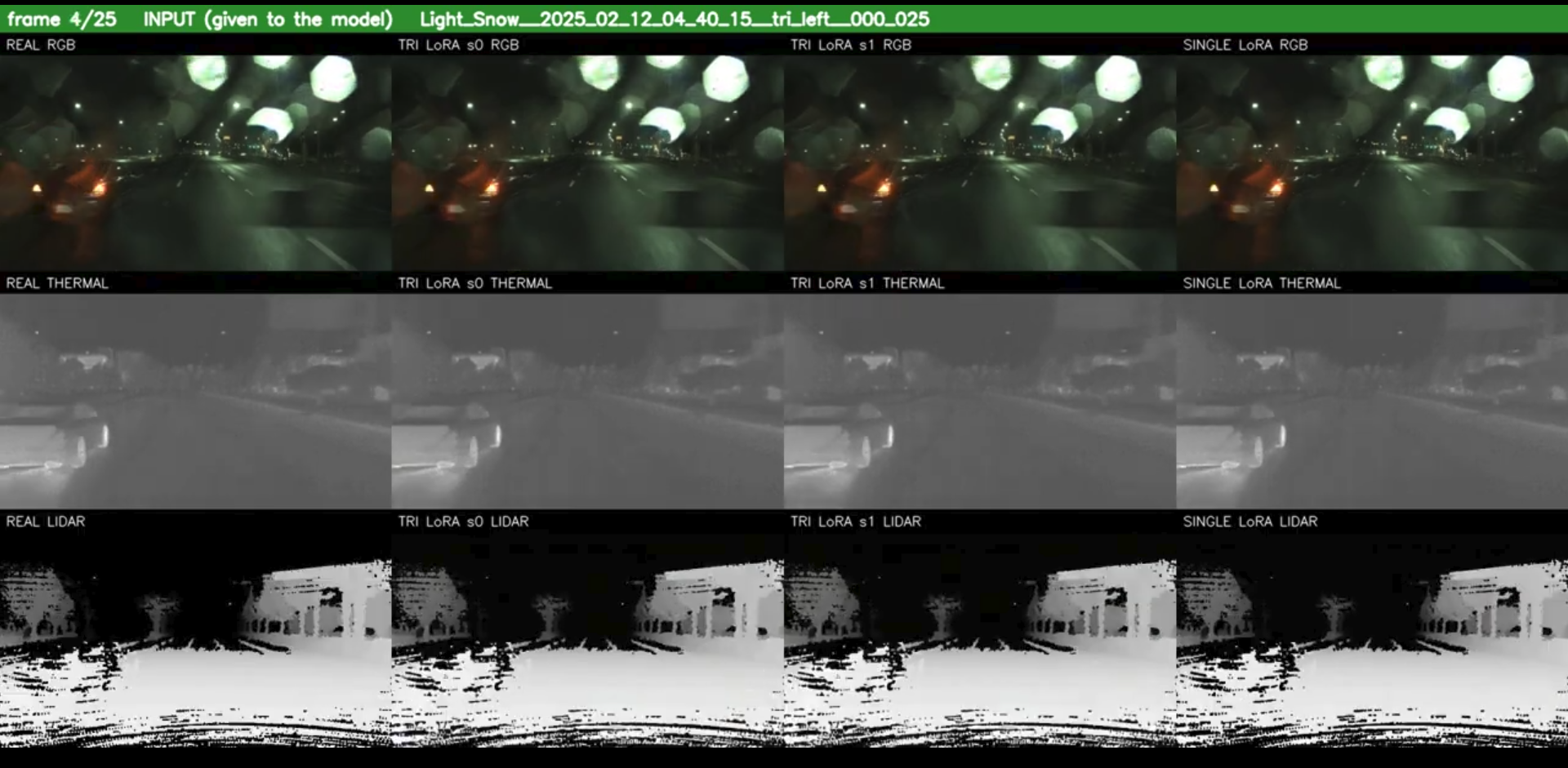}{(a) Observed frame (input)}
    \lspanel{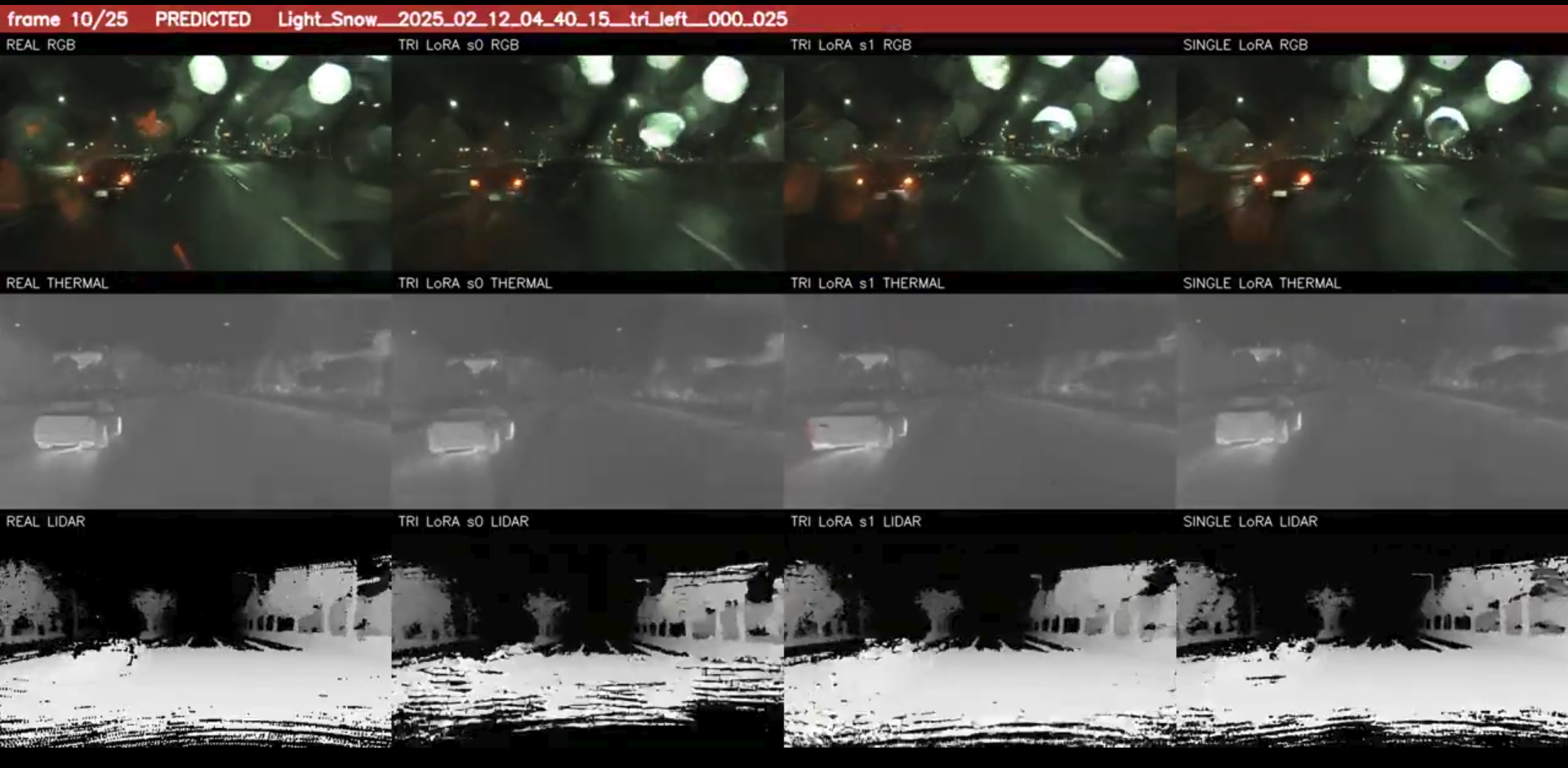}{(b) 0.5\,s after the last observed frame}
    \lspanel{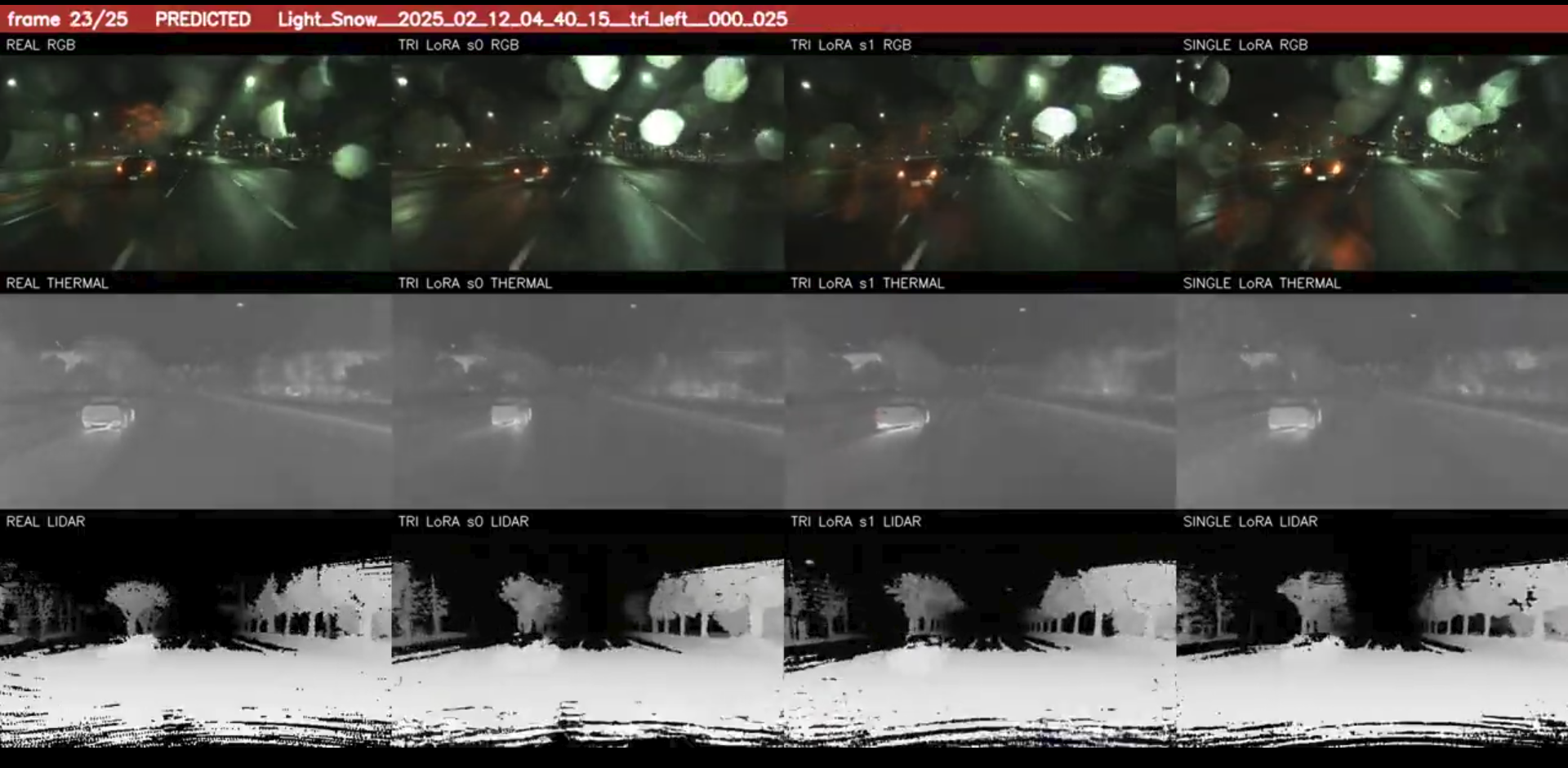}{(c) 1.8\,s after the last observed frame}
  \end{tabular}
  \caption{\textbf{A Light Snow window at night, a condition absent from training.} One sample per model; in (a), model columns show the observed input as reproduced by each model. The forecasts of the two models are very close in all three sensors and keep the car ahead and the scene layout. In (b), the tri-sensor sample leaves more gaps than the real future in the near-range LiDAR depth (bottom of the image).}
  \label{fig:lightsnow}
\end{figure}

\paragraph{Light Snow does not occur in training.}
Light Snow therefore tests a weather condition that the models have not seen. There, the RGB model has a lower LPIPS than persistence (\(-0.034\)), the thermal model gains 2.01\,dB PSNR, and the tri-sensor LiDAR forecast has a 1.34\,m lower error. These results come from five (RGB) or six (thermal, LiDAR) sequences. \Cref{fig:lightsnow} shows one Light Snow window at night: the forecasts of the tri-sensor and single-sensor models are very close to each other in all three sensors, and all of them keep the car ahead and the layout of the road and trees.

\paragraph{Joint training matters most for LiDAR in Fog.}
Fog is the condition in which the single-sensor LiDAR model fails: its error at 2\,s is 4.85\,m higher than that of compression-matched persistence (\([+1.79, +7.63]\)). In the same condition the tri-sensor model has the largest gain over the single-sensor model (\(-6.32\)\,m, \cref{sec:joint}) and is 1.46\,m better than persistence at tile resolution (\([-2.30, -0.83]\)). Across conditions, the tri-sensor LiDAR forecast beats persistence in mean error in all five conditions with at least three sequences, but in none of them at near range; the near-range difference is \(+0.67\)\,m overall and significant in Clear (\(+0.69\)\,m).

\paragraph{Heavy Snow.}
None of the 13 Heavy Snow camera windows shows a scene by our contrast test, so camera results for Heavy Snow are not reported. The LiDAR intervals for its two sequences are wide (tri-sensor minus persistence \(-2.43\)\,m, \([-7.08, +0.47]\)), and we draw no conclusion for this condition.

\subsection{Analyses on the initial split}
\label{sec:analyses}
The zero-shot comparison and the reliance analysis need runs that we made only on the initial split of 13 validation sequences (Fog 9, Light Snow 2, Heavy Snow 2; 90 windows, two samples per window).

\paragraph{Post-training versus zero-shot.}
\label{sec:zeroshot}
Sensor-specific post-training improves all three sensors over the zero-shot model (\cref{tab:zeroshot}). The gain is largest for thermal and LiDAR, the two sensors furthest from the model's RGB pretraining. The zero-shot model has never seen depth video; its LiDAR forecasts can contain structures that a depth map cannot show, and its mean error at 2\,s is twice that of persistence.

\begin{table}[t]
  \centering
  \small
  \setlength{\tabcolsep}{3.5pt}
  \caption{\textbf{Post-training helps every sensor.} Paired differences, all 90 windows of the initial split, 95\% cluster-bootstrap intervals over sequences. All intervals exclude zero.}
  \label{tab:zeroshot}
  \begin{tabular}{@{}llcc@{}}
    \toprule
    Sensor & Metric (2\,s) & LoRA \(-\) zero-shot & 95\% CI \\
    \midrule
    RGB & PSNR (dB)\,\(\uparrow\) & \(+2.17\) & \([+0.84, +3.91]\) \\
        & LPIPS\,\(\downarrow\) & \(-0.042\) & \([-0.069, -0.023]\) \\
    Thermal & PSNR (dB)\,\(\uparrow\) & \(+8.94\) & \([+5.77, +12.70]\) \\
            & LPIPS\,\(\downarrow\) & \(-0.107\) & \([-0.130, -0.084]\) \\
    LiDAR & MAE (m)\,\(\downarrow\) & \(-7.5\) & \([-9.7, -5.1]\) \\
          & \(\delta_1\)\,\(\uparrow\) & \(+0.243\) & \([+0.187, +0.289]\) \\
    \bottomrule
  \end{tabular}
\end{table}

\paragraph{Compression-matched persistence changes the verdict for thermal.}
\label{sec:fair}
Against plain persistence, the thermal model appears worse: LPIPS at 2\,s is higher by 0.014 (95\% CI \([0.007, 0.020]\)), better in only 13\% of windows. The tokenizer alone costs 0.022 LPIPS on thermal reconstructions, more than this gap. Against compression-matched persistence the difference is \(-0.002\) (\([-0.007, +0.003]\)), no clear difference, and at 1\,s the thermal model is better in PSNR on scene-visible windows (\(+1.16\)\,dB, \([+0.42, +1.80]\)). The conclusion ``the thermal model loses to persistence'' is an artifact of comparing a decoded forecast with an undecoded baseline. The full validation set shows the same pattern. In Fog, thermal LPIPS is worse than plain persistence (\(+0.017\), \([+0.014, +0.021]\)) and not clearly different from compression-matched persistence (\(+0.001\), \([-0.002, +0.005]\)); over all adverse windows the two comparisons give \(+0.009\) (\([+0.003, +0.015]\)) and \(-0.002\) (\([-0.007, +0.005]\)). For RGB and LiDAR the matching changes the values only slightly; it flips two borderline verdicts on secondary metrics (the RGB object-count error and LiDAR \(\delta_1\)) and none on the main image and depth metrics.

\paragraph{Reliance on the other sensors.}
\label{sec:reliance}
To measure how much each forecast relies on the other sensors, we blank one sensor's tile in the observed frames of the tri-sensor model and measure how much the forecasts of the other two degrade (\cref{tab:reliance}). In most cases a forecast degrades when another sensor is removed, so the tri-sensor model uses information across sensors. The exception is the LiDAR forecast, which does not change clearly when thermal is removed, nor in fog when RGB is removed. The reliance is much stronger in snow than in fog: removing RGB costs the LiDAR forecast 4.5\,m in snow and 0.3\,m in fog. Pooled over all windows, the reliance on LiDAR correlates with the number of LiDAR returns (Spearman \(\rho=0.34\) and \(0.40\) for the RGB and thermal forecasts, \(p\le0.001\)). Within fog, where the weather is fixed, this correlation disappears (\(\rho=0.09\) and \(0.12\), \(p\ge0.39\)), and so do the within-fog correlations for every other pair of blanked and forecast sensor (\(|\rho|\le0.18\), \(p\ge0.20\)). The pooled correlations therefore reflect the weather condition rather than per-window sensor quality. A black tile never occurs in training, so part of each degradation may come from the unusual input itself; the comparison between weather conditions should be less affected by this than the absolute values.

\begin{table}[t]
  \centering
  \small
  \setlength{\tabcolsep}{3pt}
  \caption{\textbf{Reliance grows from fog to snow.} Degradation of a forecast when another sensor is blanked in the observed frames (positive = worse). Tri-sensor model, initial split, one sample per blanked run. Snow pools Light and Heavy Snow (2 sequences each), so snow values are indicative. \(^\dagger\)95\% interval includes zero.}
  \label{tab:reliance}
  \begin{tabular}{@{}lllcc@{}}
    \toprule
    Blanked & Forecast & Metric & Fog (9 seq.) & Snow (4 seq.) \\
    \midrule
    RGB & LiDAR & MAE (m) & \(+0.34\)\(^\dagger\) & \(+4.47\) \\
    RGB & Thermal & LPIPS & \(+0.012\) & \(+0.020\) \\
    Thermal & RGB & LPIPS & \(+0.008\) & \(+0.089\) \\
    Thermal & LiDAR & MAE (m) & \(+0.02\)\(^\dagger\) & \(+0.20\)\(^\dagger\) \\
    LiDAR & RGB & LPIPS & \(+0.038\) & \(+0.203\) \\
    LiDAR & Thermal & LPIPS & \(+0.012\) & \(+0.028\) \\
    \bottomrule
  \end{tabular}
\end{table}

\subsection{Robustness of the evaluation protocol}
\label{sec:robust}
The full-validation protocol scores one sample and 3 of every 5 windows per sequence. We tested this choice on the initial split, where all windows and two samples are available, by recomputing every paired comparison with the first sample on all windows and on the subset. Across all comparisons with at least three sequences, no verdict reverses between better and worse, 94\% of the overall verdicts are identical, and the mean differences correlate at \(r=0.993\). Every changed verdict moves between a significant result and no clear difference. In 50 random half-subsets, no verdict reverses either.

%% file: sec/6_conclusion.tex
\section{Limitations}
\label{sec:limitations}
Our main limitation is compute. The full DSERT-RoLL release is about 268\,GB, and we train on a subset of about 30\,GB: 35 of the 132 official training sequences, without Light Snow, so Light Snow results test an unseen condition. Generation takes about 12 minutes per forecast on one A100 GPU, so the full-validation results use one sample and 3 of every 5 windows per sequence (\cref{sec:robust}), and the zero-shot comparison, the reliance analysis, and the distribution metrics are reported on the initial split only. For the same reason we study one backbone and a 2\,s horizon. Training on the full dataset and sampling more forecasts per window are left to future work.

\section{Conclusion}
\label{sec:conclusion}
We studied the joint forecasting of RGB, thermal, and LiDAR observations with a video world model in real adverse weather. A video foundation model can forecast LiDAR once the sweeps are turned into a depth video that its tokenizer can reconstruct, and adding morphological closing to the densification was the step that made this work. The persistence baseline must pass through the same tokenizer as the model: without this, the thermal model looks worse than persistence, and with it the difference disappears. Compared with the single-sensor model, the tri-sensor model lowers the LiDAR error by 2.1\,m and makes LiDAR forecasts better than compression-matched persistence, except at near range, while its camera forecasts stay close to those of the single-sensor models, with a small loss in LPIPS. A LiDAR model trained alone does not beat persistence. When one sensor is blanked, the forecasts of the others degrade more in snow than in fog, but within one weather condition the degradation does not follow the measured quality of the blanked sensor.

%% file: sec/acknowledgement.tex
\section*{LLM Usage Statement}
\label{sec:llmusagestatement}
We used LLM-based coding and writing assistants. ChatGPT (GPT-5) was used to
refine grammar, format the LaTeX of equations and tables, and condense some
sections to fit the page limit. Claude Code was used to write code for the
experiments and analysis, to check the text against the results, and to
revise the text for consistency. All experiments were run, and all results,
claims and citations were checked, by the authors, who take full
responsibility for the content.

%% file: sec/X_suppl.tex
\section{Further LiDAR depth metrics}
\begin{table}[H]
  \centering
  \small
  \setlength{\tabcolsep}{2.6pt}
  \caption{\textbf{Further LiDAR depth metrics}, full validation, 2\,s horizon. AbsRel and RMSE over all measured points; MAE (m) per distance band; hole = share of measured points whose forecast pixel is empty. Lower is better for all columns; matched = compression-matched. Bold = best at tile resolution.}
  \label{tab:depth}
  \begin{tabular}{@{}lcccccc@{}}
    \toprule
    Method & AbsRel & RMSE & Near & Mid & Far & Hole \\
    \midrule
    \multicolumn{7}{@{}l}{\emph{Full resolution}} \\
    Persistence & 0.379 & 13.65 & 3.47 & 5.56 & 12.13 & 0.058 \\
    Persist., matched & 0.379 & 13.30 & 3.45 & 5.58 & 12.14 & 0.049 \\
    Single LoRA & 0.582 & 14.22 & 7.61 & 5.73 & 10.75 & 0.042 \\
    \midrule
    \multicolumn{7}{@{}l}{\emph{Tile resolution}} \\
    Persistence & 0.349 & 12.99 & \textbf{2.99} & 5.34 & 12.13 & 0.052 \\
    Persist., matched & 0.351 & 12.28 & 3.03 & 5.36 & 12.03 & 0.039 \\
    Single LoRA & 0.550 & 13.57 & 7.12 & 5.49 & 10.78 & 0.037 \\
    Tri LoRA & \textbf{0.327} & \textbf{10.41} & 3.70 & \textbf{4.10} & \textbf{9.30} & \textbf{0.025} \\
    \bottomrule
  \end{tabular}
\end{table}

\section{Sample variability}

How much two samples for the same window differ depends strongly on the sensor. The better of two samples, chosen with the real future, improves the mean error over the average sample by 30\% for LiDAR (MAE at 2\,s), but only by 7.5\% for RGB and 3\% for thermal (LPIPS at 2\,s). For LiDAR, the failure mode is a forecast that fades toward mid-gray, \ie toward a constant depth of about 40\,m. This failure can be detected without the real future: keeping the sample with the higher spread of depth codes recovers 82\% of the gain of the oracle choice (90\% and 71\% on two disjoint halves of the sequences), and makes the single-sensor LiDAR model better than compression-matched persistence at 1\,s (MAE \(-1.76\)\,m, \([-2.78, -0.90]\)). The selection rule was found on this split, so we report it as an analysis rather than as part of the method.